\documentclass[runningheads]{llncs}
\usepackage{doi}
\usepackage{graphicx}
\usepackage{subfigure}
\usepackage{appendix}

\usepackage{amsthm}
\usepackage{amsmath}
\usepackage{amssymb}
\usepackage[mathscr]{euscript}
\usepackage{array}

\renewcommand{\doi}[1]{\href{https://doi.org/#1}{doi:#1}}

\usepackage{url}

\usepackage[T1]{fontenc} 
\usepackage{setspace} 

\usepackage{color}
\usepackage{xcolor}
\usepackage{multicol}
\usepackage{multirow}
\usepackage{hhline}

\usepackage{xcolor}

\begin{document}
\title{Teaching MLP More Graph Information: A Three-stage Multitask Knowledge Distillation Framework.\thanks{Supported by organization x.}}
%
%
\author{
Junxian Li\inst{1} \and 
Bin Shi\inst{1} \and 
Erfei Cui\inst{1} \and 
Hua Wei \inst{2} \and
Qinghua Zheng\inst{3}}
%
%
\institute{Xi'an Jiaotong University \and
Arizona State University, Tempe, USA \and
Tongji University, Shanghai, China
}

%
\maketitle              
\begin{abstract}
We study the challenging problem for inference tasks on large-scale graph datasets of Graph Neural Networks: huge time and memory consumption, and try to overcome it by reducing reliance on graph structure. Even though distilling graph knowledge to student MLP is an excellent idea, it faces two major problems of positional information loss and low generalization. To solve the problems, we propose a new three-stage multitask distillation framework. In detail, we use Positional Encoding to capture positional information. Also, we introduce Neural Heat Kernels responsible for graph data processing in GNN and utilize hidden layer outputs matching for better performance of student MLP's hidden layers. To the best of our knowledge, it is the first work to include hidden layer distillation for student MLP on graphs and to combine graph Positional Encoding with MLP. We test its performance and robustness with several settings and draw the conclusion that our work can outperform well with good stability. Code implementation is in \texorpdfstring{\url{https://github.com/lijunxian111/KMP}}{https://github.com/lijunxian111/KMP}.

\keywords{Knowledge distillation on graphs  \and Positional encoding \and Hidden layer distillation.}
\end{abstract}

\section{Introduction}

Graph Neural Networks (GNNs)\cite{ref_article1} have shown superior performance in learning structural information and are widely used in various fields, such as molecular biology, social networks, and recommendation systems\cite{ref_article3,ref_lncs3,ref_article4}. With the continuous growth of applications and data scale, GNN tasks exhibit the characteristics of big models and big data. For instance, a frequently used Twitter-7~\cite{ref_lncs4} dataset contains over seventeen million nodes and over four hundred million edges.

However, this trend poses a major drawback: the required computing resources are sharply increased. It is difficult to train conventional GNNs on large-scale datasets with limited resources and training time. Therefore, it is essential to propose efficient approaches for large-scale graph deep learning with acceptable performance.

A large number of efforts have been made to tackle this issue in both hardware-level and algorithm-level~\cite{ref_lncs5,ref_article5}. For example, some studies optimize the cache and GPU usage, while others speed up and enhance graph data computation by sampling, clustering~\cite{ref_lncs6}, quantization~\cite{ref_lncs7}, and pruning~\cite{ref_article6}. Nevertheless, these methods still retain the graph aggregation process that consumes resources. Moreover, the prediction performance is poor on unseen classes. To address these issues, recent works have tried to use knowledge distillation on graphs~\cite{ref_lncs8,ref_article7,ref_article8}. Among them, GLNN~\cite{ref_article8} considers MLP, which is frequently used in the industrial area, as a student and transfers knowledge into it by matching its soft logits with the soft target distribution from the teacher. MLP serves as a simple, clear model and can perform well with only node features for inference.

Despite the efficiency of GLNN, there are problems with this type of knowledge distillation, which will restrict it to reach optimal prediction accuracy:

\textbf{1) Positional information loss problem.} Using student MLP as a classifier relies on the assumption that it is sufficient to perform classification with only node features. However, this is unsuitable for graph datasets where positional information is highly correlated with classification results. A simple MLP with raw node features as inputs cannot make use of this information.
\textbf{2) Low generalization problem.} GLNN uses soft logits from the teacher GNN to teach student MLP the mapping relationships between node features and labels. However, the relationships learned by MLP are usually too simple and direct, leading to low generalization probability.

In this paper, we propose a new multi-stage framework for knowledge distillation on graphs that efficiently solves the problems above. To address the positional information loss problem, we introduce a lightweight Laplacian Positional Encoding that captures the graph's positional information and concatenates it with the initial node feature. This approach improves the performance and applicability of student MLP on different datasets. To enhance the generalization capability of the distillation, we draw inspiration from the GNN-level topology distillation~\cite{ref_lncs9,ref_article9} and FitNet~\cite{ref_article10}. We teach the student the process of message propagation in the teacher GNN through a special hidden layer distillation. By mapping the student's and teacher's hidden layer outputs to the same space using kernel functions, we minimize the distance between the transformed results. This approach ensures that the student MLP captures the information responsible for the message-passing process, leading to better node representations of hidden layers and enhanced generalization. We validate our framework's superiority experimentally and investigate its robustness. Our experiments demonstrate that the proposed framework outperforms existing state-of-the-art methods on various benchmark datasets.

The main contributions of this work are:

---Introducing Laplacian Positional Encoding as an initial feature for student MLP to capture the positional information of graphs and improve its performance and applicability on different datasets.

---Utilizing special hidden layer distillation to teach the student the process of message propagation in the teacher GNN and improve its generalization capability.

---Conducting experiments on both small and large scale data, showing that our approach significantly outperforms existing methods. We also have interesting cases to illustrate the robustness of our methods.

Overall, our framework provides a novel approach to knowledge distillation on graphs that is both efficient and effective.

\section{Related Work}
\textbf{Graph Neural Networks. }GNNs are primarily used to analyze and study data in non-Euclidean domains that can be represented as graphs $G(V,E)$. These models allow for the incorporation of two types of knowledge: node and structural information. Most commonly used GNNs, for instance, GCN~\cite{ref_lncs1}, GAT~\cite{ref_article11} and GraphSAGE~\cite{ref_article12}, are all based on message-passing neural networks (MPNN) represented as follows:
\begin{equation}
    X_s^{l+1}=f(X_s^l, A ggregate(MSG(X_s^l,X_{n_j}^l,edge_{sn_j})))
\end{equation}
where $X_s$ means each node itself and $X_{n_j}$ means certain node from neighborhood node set. The process of aggregating and computing functions is highly dependent on the topology structure of the graph in order to achieve optimal performance.

\textbf{Model-level GNN Acceleration. } 
The field of GNN acceleration addresses two major challenges. The first challenge is that \textbf{large-scale graph datasets} used in real-world applications make GNN training and inference difficult to complete within a reasonable time frame. The second challenge arises from \textbf{the development of more complex and deeper GNN models}, such as DeeperGCN~\cite{ref_article13} and Graph Transformer~\cite{ref_article14}, which require extensive training time that is almost impossible to complete without time-complexity optimization.
To overcome these challenges, researchers have proposed model-level optimization techniques. GNN simplification methods, such as SGC~\cite{ref_lncs10} and LightGCN~\cite{ref_lncs11}, aim to simplify the computational process in GNNs. On the other hand, GNN compression techniques, such as pruning, model quantization, and knowledge distillation, aim to replace cumbersome primitive models with lightweight models that require fewer parameters. By employing these techniques, researchers hope to reduce the time and computational resources required for GNN training and inference, making it feasible to apply GNNs to large-scale graph datasets and more complex models.

\textbf{Knowledge Distillation for Graphs. }Knowledge distillation methods have been proposed to transfer knowledge from complex teacher models to simpler student models. However, for graph-based models, the task is a bit challenging. Localized graph-based methods, such as LSP~\cite{ref_lncs12} and TinyGNN~\cite{ref_lncs13}, focus on capturing local information of graphs. GraphAKD~\cite{ref_lncs14} proposes an adversarial framework for knowledge distillation that allows students to generate node embeddings and logits similar to their teachers' output. All these methods highly rely on graphs for knowledge distillation.
In contrast, GLNN, as mentioned above, aims to transfer knowledge into a simpler student MLP but does not utilize structural knowledge. It is effective at knowledge transferring to simpler models. However, a further optimization can be introduced. 

\section{Preliminaries}

\subsection{Motivation. }
To improve knowledge transferring to student MLPs, it's advisable to use distillation methods that go beyond just aligning soft label distributions. By defining the graph signal $X$, we consider two species of knowledge: \emph{local node representation and global positional information.} Here we introduce approach called \textbf{Neural Heat Kernel(NHK) Based Distillation} for effective node representation generation from student MLPs. Moreover, we study \textbf{Positional Encoding} (PE) for better global positional information transfer. 
\subsection{NHK Based Distillation} \label{Sec:3.2}
To introduce NHK as a teaching tool for MLPs, it's essential to include the following definition:
\begin{definition}
\textbf{GNN from Riemannian manifold perspective. }
In this perspective, the graph data can be defined as a special kind of Riemannian manifold. When graph signal $X_{sig}$ defined, a single-layer GNN is equivalent to solving the heat equation for a discrete value of some time interval $t$. It can also be written with a kernel function $f_k$:
\begin{equation}
    \frac{\partial u(X_{sig},t)}{\partial t}=-c\Delta(X_{sig},t)
\end{equation}
where $X$ is the spatial signal, also graph structure we have. And $t$ is the temporal signal related to the number of GNN layers. $\Delta$ denotes for the Laplace operator. Usually we consider kernel functions and distance measuring to
approximate this operator. Kernel functions will be discussed in Sec.~\ref{section:4}.
\end{definition}
This transformation can facilitate the extraction of local structural features. We resort to the NHK to capture the influence exerted on the nodes by the message passing process during the graph convolution process, for the process of information exchange between two points on the manifold is similar to that of heat transform. Consequently, feature embedding of nodes after certain hidden layers is added with information from connected nodes with the mapping from kernel functions $f_k$. This can be regarded as a prior knowledge also named \textbf{Graph Smoothness}~\cite{ref_lncs15}. Layer-to-layer kernel mapping are proposed as:
\begin{equation}
    H_{\alpha}^{(l+a)}=\sum \frac{f_k^{(l \ to \ l+a)}(node_{\alpha},node_{\beta}) H_{\beta}^{(l)}}{degree_{\beta}} \quad (a \in 1,2,...,n)
\end{equation}
where node-level smoothness is passed between $node_\alpha$ and $node_\beta$, which are arbitrary nodes from graph. $H_{\alpha}$ and $H_{\beta}$ are node representations after certain hidden layer. By characterizing fuzzy geometric information as a specific function, we adopt an easily comprehensible approach that simplifies local distillation, which will be discussed further below. 

\subsection{Positional Encoding for global distillation. } To improve the capturing of global information, we draw inspiration from graph transformer frameworks. Specially, Dwivedi et al.~\cite{ref_article16} utilize Laplacian Encoding as initial PE, calculated before training. By providing PE as prior knowledge to student model, its ability to perceive positional and structural information can be enhanced. In this work, we use Laplacian eigenvectors as PE. The eigenvectors are calculated by defining a Laplacian matrix:
\begin{equation}
    \Delta = I-D^{-\frac{1}{2}}AD^{-\frac{1}{2}}=U^T{\Lambda}U
\end{equation}
where $A$ is the adjacency matrix, $D$ is the diagonal matrix for matrix standardization and $U$, $\Lambda$ represents eigenvectors and eigenvalues of the Laplacian matrix. Then we choose \textbf{k-smallest} non-trivial eigenvectors in real space for n nodes as an extra input $U_{selected}\in R^{n\times k}$. A theorem related to Spectral Clustering~\cite{ref_article15} demonstrates benefits of PE for classification tasks: 
\begin{theorem}
\textbf{(Effectiveness of Spectral Clustering).} Suppose that whole graph consists of multiple graph partitions that are similar internally and differ significantly between partitions. Thus, we need to find such optimal partitions. Given a partition of the graph into k sets, we can define k indicator vectors $H=(h_1,h_2,...,h_k)$ by approximating Ratiocut or Ncut. We consider the problem of finding ideal indicator vectors transforming into the following Ncut minimizing problem:
\begin{equation}
    \min Ncut(A_1,A_2,...,A_k)
\end{equation}
It can be proven that clustering can provide structural information by providing similarities and Euclidean distances as prior knowledge. According to Perturbation theory, the eigenvectors of Laplacian matrices will be very similar to the ideal indicator vectors. Therefore, the first-k eigenvectors are often used in clustering.
\end{theorem}

\subsection{KL-Divergence} To quantify the degree of difference in the soft target distribution between student and teacher models, we introduce Kullback-Leible Divergence~\cite{ref_lncs16} (KL divergence). Consider two random variables $P$, $Q$, with discrete probability distributions(like the real situations) $p(\theta)$ and $q(\theta)$, respectively. We have KL divergence from P to Q as:
\begin{equation}
    D_{KL}(p||q)= \sum_{i=1}^n p(\theta)\log \frac{p(\theta)}{q(\theta)}
\end{equation}
The smaller the difference between the soft logits of the student and teacher models, the smaller the KL divergence will be. This shows that we can use minimizing KL divergence as a way to simplify the utilization of soft target information.
\section{Methodology}
\label{section:4}
\begin{figure}[t]
\includegraphics[width=\textwidth]{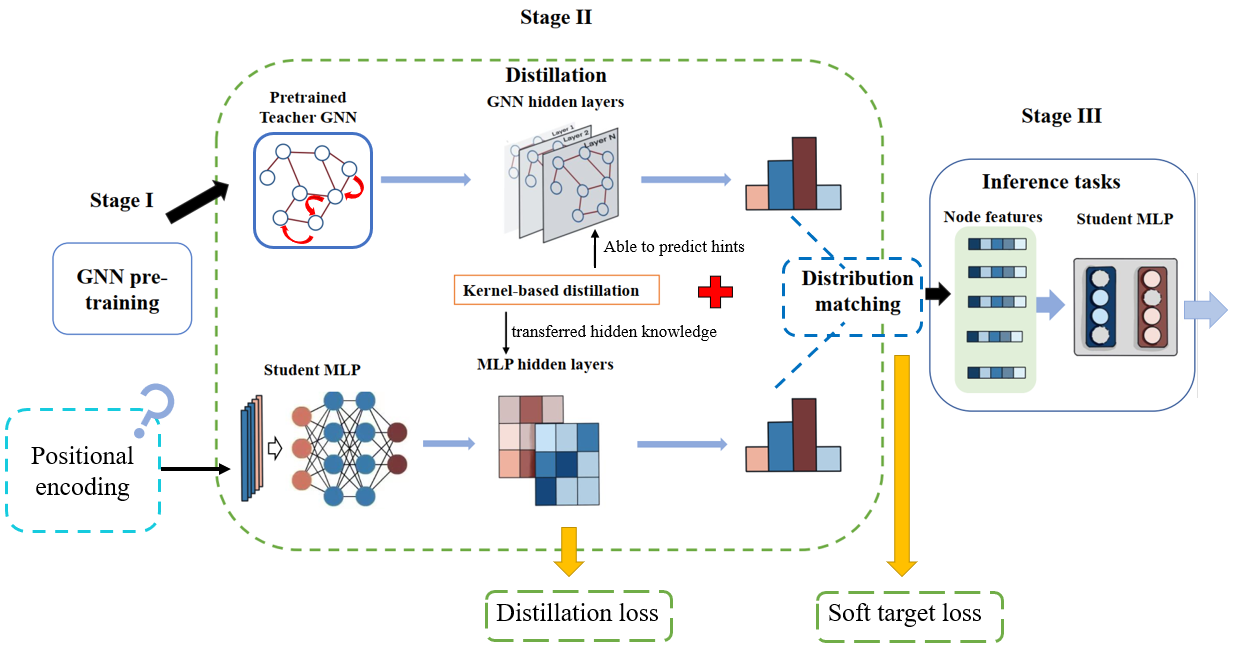}
\caption{Overall structure of our framework. It consists of three stages: GNN pretraining, distillation and inference. We transfer knowledge to student MLP during stage II. Best viewed in color.}
\label{fig1}
\end{figure}

To transfer graph information to MLP effectively, it's necessary to consider both local and global perspect. Combining the above methods introduced, we propose a multi-task distillation framework \textbf{K}ernel-based \textbf{M}ultilayer Perceptron with structural \textbf{P}rocessing (KMP) for better results of teaching student MLPs. The overall structure is shown in Fig.~\ref{fig1}.

\subsection{Three-Stage distillation }
\label{Sec:4.1}
Our proposed distillation process consists of three stages: GNN pre-training stage, distillation stage and inference stage. During the first stage, we perform simple message-passing training and save the trained graph model $GNN_{pre}$ and the soft target distribution $T_t$ of the model output. 

During the second stage, we first add PE as a prior knowledge for student MLPs. Consider the $d$-dimensional input feature $X_{s_i}^{(0)} \in R^d$ of a certain node. We embed the $k$-dimensional Laplacian PE into a feature vector with the same dimension as $X_{s_i}^{(0)}$. Then we concatenate them or do a simple dot product between them. As the following equation shows:
\begin{equation}        
x_{pos}^{(0)}=Emb_{p}(PE_{i})=K^{0}PE_{i}+b^{0} \in R^{d}, 
\quad K^{0} \in R^{k\times d}, b^{0} \in R^d
\end{equation}
\begin{equation}
    X_{s_i}^{(0)}=concat[X_{s_i}^{(0)};x_{pos}^{(0)}] \quad or \quad mul(X_{s_i}^{(0)};x_{pos}^{(0)})
\end{equation}
In order to capture the $message-passing \ control \ information$, we make sure that nodes for training contain the same ones as being input towards teacher GNNs. Thus, the process of hidden mechanism control by \textbf{Neural Heat Kernels} can work. Our target is to make hidden layers of students able to predict the outputs of a guiding hidden layer of teachers~\cite{ref_article10}. Layer-to-layer mapping matrices as Sec.~\ref{Sec:3.2} on hidden outputs of $m$ training nodes are adopted as:
\begin{equation}
Mat=\mathbb{M}[f_k^{(l \ to \ l+a)}(h_{\alpha},h_{\beta})]\in R^{m \times m}
\end{equation}
where $\mathbb{M}[\bullet]$ denotes for operation to obtain mapping matrices. We select four non-linear kernel functions as $f_k^{(l \ to \ l+a)}$ to do the calculation:
\begin{equation}
 \begin{cases}
 \sigma(f_{proj}(\langle h_{\alpha}^{(l)},h_{\beta}^{(l)} \rangle)) \quad \quad \quad \quad \quad \quad (Sigmod \ Kernel) \\
 \frac{1}{t}\sum_{r=1}^te^{\xi_{r}}\sigma(M_rh_{\alpha})^T\sigma(M_rh_{\beta}) \quad (Randomized \ Kernel) \\
 (h_{\alpha}^{(l)},h_{\beta}^{(l)}+c)^d \quad\quad\quad\quad\quad\quad\quad(Polynomial \ Kernel) \\
 e^{-\frac{||h_{\alpha}^{(l)}-h_{\beta}^{(l)}||_2^2}{4\times T}}, \quad \quad \quad \quad \quad \quad \quad \quad(Gaussian\ Kernel)
 \end{cases}
\end{equation}
where $\sigma$ denotes for non-linear operations like $Sigmod$, $Tanh$ or $Relu$ ,$f_{proj}$ denotes for a linear transformation in Sigmod Kernel and $T$ denotes for a constant time interval in Gaussian Kernel. And $M_r$ is a randomized matrix whose vectors obey a Gaussian distribution in Randomized Kernel. We load the pre-trained $GNN_{pre}$ in the first stage to generate teacher mapping matrices and aim to minimize the distance between those of teachers and students. A simple L2 Norm is used to measure distance as:
\begin{equation}
dis(Mat_s^{(l)},Mat_t^{(l)})=||Mat_s^{(l)},Mat_t^{(l)}||^2_2
\end{equation}
Also, we adopt a process of soft logits matching, which is defined in Section~\ref{Sec:4.2}. During stage III, we do inference tasks on either the existing graph or unseen new in the same graph dataset. We will discuss this further in Sec.~\ref{Sec:5.1}.
\subsection{Distillation loss } \label{Sec:4.2}
During the entire training process, we consider two main parts of the distillation loss: the loss from model prediction matching and the loss from distance measurement between mapping matrices. Therefore, we can express the total multi-task distillation loss $L_{total}$ as:
\begin{equation}
\mathcal{L}_{total}=\mathcal{L}_{out}(\hat{Y_s},Y_t)+\frac{\gamma}{K}\sum_{l=1}^K\mathcal{L}_{dis}(Mat_s^{(l)},Mat_t^{(l)})
\end{equation}
where $\hat{Y_s}$ denotes for student prediction, and $Y_t$ can be true labels or soft targets from teacher. $\gamma$ is a constant to show the proportion of $\mathcal{L}_{dis}$. And that is calculated by a K-layer matching process between $Mat_s^{(l)}$ and $Mat_t^{(l)}$ included above. It is noteworthy that we provide two kinds of training nodes: with true labels or with soft targets from teacher GNNs. The student predictions of them can be represented as $\hat{Y_{sL}}$ and $\hat{Y_{sO}}$. So $\mathcal{L}_{out}$ can be specified as:
\begin{equation}
\mathcal{L}_{out}=\theta\mathcal{L}_{truth}(\hat{Y_{sL}},y_t)+(1-\theta)\mathcal{L}_{soft}(\hat{Y_{sO}},T_t), \quad \theta \in [0,1]
\end{equation}
where $y_t$ are true labels of selected labeled nodes while $T_t$ are soft target distributions from teacher outputs, as mentioned in Sec.~\ref{Sec:4.1}. They are two parts of $Y_t$. By minimizing $\mathcal{L}_{total}$, we can get a well-trained MLP to do inference tasks.
\subsection{Usage of trainable reverse kernel} \label{sec:4.3}
Getting inspired by Variational Autoencoder~(VAE)~\cite{ref_article22} and GNN-level topology distillation~\cite{ref_article9}, We have tried to optimize the heat kernel itself. A trainable matrix $W_k$ is introduced for transformation whose parameters can be optimized and shared by both student and teacher. The trainable reverse kernel can be represented as:
\begin{equation}
f_k^{(l+a\ to \ l)}=\sigma(W_kh_{\alpha}^{(l+a)})^T \sigma(W_kh_{\beta}^{(l+a)})
\end{equation}
Like VAE does, we also adopt reconstruction loss between original inputs($X^{(0)}$) and outputs of the last hidden layer ($H^{(l+a)}$)  and utilize it to optimize $W_k$ for data processing:
\begin{equation}
    \mathcal{L}_{re}=||f_k^{(l+a\ to \ l)}H^{(l+a)},X^{(0)}||^2_2
\end{equation}
Consequently, we need to do a two-step optimization both on the student model and the kernel. That helps to explore proper kernels for distillation.
\subsection{Discussion}
Our proposed methodology is mainly for solving two major problems of MLP classifier: $1.$ positional information loss $2.$ low generalization. To address these issues, we utilize two efficient approaches: PE and NHK based distillation. These approaches enable the transfer of knowledge from non-Euclidean domains into simpler domains in a explainable manner. Overall, our methodology offers a promising solution to enhance the performance of MLP classifiers on graph datasets. Also, this transfer is done in a pretty explainable way. 
\section{Experiments}
In this section, we introduce several groups of experiments to prove that our \textbf{KMP} framework can solve the problems of MLP classifier. Mainly, we focus on the following questions to design experiments:
\textbf{Q1: }\emph{Does KMP perform better than student MLP without hidden layer distillation on most common node classification tasks?}

\textbf{Q2: }\emph{Can KMP give better inference results on large-scale graph datasets?}

\textbf{Q3: }\emph{Does trainable kernels work better?}

\textbf{Q4: }\emph{How much can PE improve our model performance?}

\textbf{Q5: }\emph{How is the robustness and sensitiveness of our framework?}
Before introducing our experiment results, we have the following preliminary statement: We study on question 1, 2 and 3 with a single KMP without PE, and adopt KMP+PE in question 4 and 5.
\subsection{Dataset} 
In this part, we do our research on five typical datasets~\cite{ref_article17}. Three common benchmark  datasets: Cora, Citeseer, Pubmed and two larger ones: Amazon photos, Amazon Computers. Also, we introduce two OGB datasets~\cite{ref_lncs17}: Arxiv and Products, to see if our model perform well on large-scale graphs. The details of the datasets are in Appendix.~\ref{tab:CPF_detail}. For each dataset, we label 20 nodes per class and select 30 nodes randomly for validation. We adopt both true or soft labels here. Other nodes are provided for testing. 
\subsection{Baselines} 
We choose frequently used teacher models: SAGE, GCN and GAT. It's easy for us to transfer knowledge from them into student MLPs. We use single MLP without distillation and GLNN as baseline model to check if our framework perform better. 
\subsection{Evaluation Metrics} 
In all experiments, we report accuracy($\%$)on test set. \textbf{10 experiments with random seeds are done, and mean value and standard deviation of \textbf{test accuracy} are reported.} 
\subsection{Hyperparameters} \label{Sec:5.4}
During stage I and II, we set the max epochs during training to 1000, and an early stop is adopted when accuracy on validation set doesn't increase for 50 epochs. We take 512 as our batch-size, $Adam$ as our optimizer and cuda as our training device. In order to report exact results, we follow the settings of teacher GNNs for 5 small datasets in Appendix.~\ref{appendix:settings}. A grid-research is also adopted to find the most proper hyperparameters for loss computing, as introduced in Appendix.~\ref{appendix:grid_search}. 
\subsection{Performance on most common node classification tasks} \label{Sec:5.1}
\textbf{Transductive and inductive settings. }We study two categories of node classification tasks. The most common setting for node classification is a
\textbf{transductive} (trans) setting, where all nodes are split into training set, validation set and test set. Under this setting we can see the whole graph structure (adjacency matrix) while training. However, under real-life scenarios, an \textbf{inductive} (induc) setting is often more applicable, for we cannot always see all users and relations in real networks. Unobserved nodes are first taken out and all information of them can't be seen during stage II. Then a train-valid-test split is adopted. Something to note while transductive or inductive distillation is as follows:

\noindent $\bullet$ During stage II, we choose some of the training nodes to own true labels, and others to match teacher's soft target distribution, as Sec.~\ref{Sec:4.2} shows.

\noindent $\bullet$ During stage III, we use the structure-observed test set  to do inference via our trained student MLP for transductive setting, while unobserved new nodes are used for inductive setting. 

\noindent \textbf{Results. }Overall performance of the most common node classification task can be found in Table.~\ref{tab:result_1} and Table.~\ref{tab:result_2}. For KMP, We report the best result among four categories of kernels as Sec.~\ref{Sec:4.1} mentions. Our comparison is among MLP without distillation, GLNN and our KMP (no Position Encoding here). It clearly demonstrates that our KMP outperform baseline GLNN well in nearly all cases with transductive settings. Moreover, test accuracy from KMP's predictions clearly increases from GLNN's with more discursive inductive settings on most occasions.  We can safely draw the conclusion that our distillation method can improve student MLP's performance with various teacher GNNs. 
\begin{table}[htbp]
\renewcommand\arraystretch{2}
\centering
\caption{Results of teachers, baseline models and KMP for \textbf{trans} setting. The \textbf{best} performance of student model is highlighted. }
\label{tab:result_1}
\resizebox{1\textwidth}{!}{
\begin{tabular}{|c|cccccccccc|} \hline
\multirow{2}{*}{\textbf{Dataset}} &
   \multicolumn{10}{c|}{\textbf{Models}} \\ \cline{2-11}
   &
  \multicolumn{1}{c|}{\textbf{$SAGE$}} &
  \multicolumn{1}{c|}{\textbf{$GCN$}} &
  \multicolumn{1}{c|}{\textbf{$GAT$}} &
  \multicolumn{1}{c|}{\textbf{$MLP$}} &
  \multicolumn{1}{c|}{\textbf{$GLNN(SAGE)$}} &
  \multicolumn{1}{c|}{\textbf{$KMP(SAGE)$}}
  & \multicolumn{1}{c|}{\textbf{$GLNN(GCN)$}} &
  \multicolumn{1}{c|}{\textbf{$KMP(GCN)$}} &
  \multicolumn{1}{c|}{\textbf{$GLNN(GAT)$}} &
  \textbf{$KMP(GAT)$} \\ \hline
   \textbf{Cora} &
   \multicolumn{1}{c|}{79.70$\pm$1.90} &
  \multicolumn{1}{c|}{79.58$\pm$1.01} &
  \multicolumn{1}{c|}{81.41$\pm$1.15} &
  \multicolumn{1}{c|}{58.66$\pm$2.34} &
  \multicolumn{1}{c|}{78.99$\pm$1.90} &
  \multicolumn{1}{c|}{\textbf{79.48$\pm$1.48}} &
  \multicolumn{1}{c|}{79.43$\pm$0.82} & 
  \multicolumn{1}{c|}{\textbf{80.03$\pm$1.19}} &
  \multicolumn{1}{c|}{79.89$\pm$1.44} &
  \textbf{80.44$\pm$1.13} \\
  \textbf{Citeseer} &
  \multicolumn{1}{c|}{70.53$\pm$2.16} &
  \multicolumn{1}{c|}{70.09$\pm$2.23} &
  \multicolumn{1}{c|}{71.76$\pm$1.54} &
  \multicolumn{1}{c|}{58.01$\pm$0.96} &
  \multicolumn{1}{c|}{71.23$\pm$2.13} & 
  \multicolumn{1}{c|}{\textbf{71.96$\pm$2.25}} &
  \multicolumn{1}{c|}{71.24$\pm$2.98} &
  \multicolumn{1}{c|}{\textbf{72.18$\pm$2.23}} &
  \multicolumn{1}{c|}{72.44$\pm$1.63} &
  \textbf{72.72$\pm$1.18} \\
  \textbf{Pubmed} &
  \multicolumn{1}{c|}{75.53$\pm$2.39} &
  \multicolumn{1}{c|}{77.96$\pm$2.96} &
  \multicolumn{1}{c|}{78.30$\pm$2.44} &
  \multicolumn{1}{c|}{67.72$\pm$3.34} &
  \multicolumn{1}{c|}{76.32$\pm$2.67} & 
  \multicolumn{1}{c|}{\textbf{76.80$\pm$2.01}} &
  \multicolumn{1}{c|}{77.94$\pm$2.95} &
  \multicolumn{1}{c|}{\textbf{78.48$\pm$1.90}} &
  \multicolumn{1}{c|}{76.44$\pm$2.28} &
  \textbf{76.86$\pm$2.40} \\
  \textbf{Amazon Photo} &
  \multicolumn{1}{c|}{90.26$\pm$1.14} &
  \multicolumn{1}{c|}{87.18$\pm$4.05} &
  \multicolumn{1}{c|}{91.61$\pm$0.94} &
  \multicolumn{1}{c|}{78.39$\pm$1.71} &
  \multicolumn{1}{c|}{92.04$\pm$0.91} & 
  \multicolumn{1}{c|}{\textbf{92.23$\pm$0.85}} &
  \multicolumn{1}{c|}{89.97$\pm$3.08} &
  \multicolumn{1}{c|}{\textbf{90.48$\pm$2.55}} &
  \multicolumn{1}{c|}{93.01$\pm$0.94} &
  \textbf{93.19$\pm$0.78} \\
  \textbf{Amazon Computer} &
  \multicolumn{1}{c|}{83.23$\pm$1.49} &
  \multicolumn{1}{c|}{80.19$\pm$5.18} &
  \multicolumn{1}{c|}{83.03$\pm$2.30} &
  \multicolumn{1}{c|}{69.01$\pm$3.27} & 
  \multicolumn{1}{c|}{84.41$\pm$1.51} &
  \multicolumn{1}{c|}{\textbf{84.75$\pm$1.10}} &
  \multicolumn{1}{c|}{78.42$\pm$4.26} &
  \multicolumn{1}{c|}{\textbf{78.96$\pm$4.27}} &
  \multicolumn{1}{c|}{\textbf{83.52$\pm$2.17}} &
  82.94$\pm$2.00 \\ \hline
\end{tabular}
}
\vspace{-3mm}  
\end{table}

\begin{table}[htbp]
\renewcommand\arraystretch{2}
\centering
\caption{Results of teachers, baseline models and KMP for \textbf{induc} setting. The \textbf{best} performance of student model is highlighted. }
\label{tab:result_2}
\resizebox{1\textwidth}{!}{
\begin{tabular}{|c|cccccccccc|} \hline
\multirow{2}{*}{\textbf{Dataset}} &
   \multicolumn{10}{c|}{\textbf{Models}} \\ \cline{2-11}
   &
   \multicolumn{1}{c|}{\textbf{$SAGE$}} &
  \multicolumn{1}{c|}{\textbf{$GCN$}} &
  \multicolumn{1}{c|}{\textbf{$GAT$}} &
  \multicolumn{1}{c|}{\textbf{$MLP$}} &
  \multicolumn{1}{c|}{\textbf{$GLNN(SAGE)$}} &
  \multicolumn{1}{c|}{\textbf{$KMP(SAGE)$}} &
  \multicolumn{1}{c|}{\textbf{$GLNN(GCN)$}} &
  \multicolumn{1}{c|}{\textbf{$KMP(GCN)$}} &
  \multicolumn{1}{c|}{\textbf{$GLNN(GAT)$}} &
  \textbf{$KMP(GAT)$} \\ \hline
   \textbf{Cora} &
  \multicolumn{1}{c|}{78.64$\pm$2.06} &
  \multicolumn{1}{c|}{78.90$\pm$1.69} &
  \multicolumn{1}{c|}{81.10$\pm$2.46} &
  \multicolumn{1}{c|}{58.02$\pm$2.72} &
  \multicolumn{1}{c|}{\textbf{71.94$\pm$2.85}} & 
  \multicolumn{1}{c|}{71.85$\pm$2.40} &
  \multicolumn{1}{c|}{72.39$\pm$2.30} &
  \multicolumn{1}{c|}{\textbf{72.86$\pm$2.37}} &
  \multicolumn{1}{c|}{72.41$\pm$1.48} &
  \textbf{72.53$\pm$1.38} \\
  \textbf{Citeseer} &
  \multicolumn{1}{c|}{70.83$\pm$2.83} &
  \multicolumn{1}{c|}{68.26$\pm$3.65} &
  \multicolumn{1}{c|}{70.75$\pm$3.29} &
  \multicolumn{1}{c|}{59.46$\pm$4.55} &
  \multicolumn{1}{c|}{70.80$\pm$2.63} &
  \multicolumn{1}{c|}{\textbf{71.22$\pm$2.04}} & 
  \multicolumn{1}{c|}{67.15$\pm$2.33} &
  \multicolumn{1}{c|}{\textbf{68.04$\pm$2.52}} &
  \multicolumn{1}{c|}{68.92$\pm$2.94} &
  \textbf{69.17$\pm$3.01} \\
  \textbf{Pubmed} &
  \multicolumn{1}{c|}{74.86$\pm$2.97} &
  \multicolumn{1}{c|}{74.40$\pm$2.94} &
  \multicolumn{1}{c|}{76.64$\pm$2.97} &
  \multicolumn{1}{c|}{66.14$\pm$5.03} &
  \multicolumn{1}{c|}{74.76$\pm$2.69} &
  \multicolumn{1}{c|}{\textbf{74.92$\pm$3.03}} & 
  \multicolumn{1}{c|}{74.89$\pm$2.81} &
  \multicolumn{1}{c|}{\textbf{75.44$\pm$2.75}} &
  \multicolumn{1}{c|}{\textbf{74.63$\pm$2.81}} &
  74.53$\pm$3.06 \\
  \textbf{Amazon Photo} &
  \multicolumn{1}{c|}{91.33$\pm$1.56} &
  \multicolumn{1}{c|}{89.28$\pm$2.12} &
  \multicolumn{1}{c|}{92.27$\pm$1.50} &
  \multicolumn{1}{c|}{79.01$\pm$1.80} &
  \multicolumn{1}{c|}{89.75$\pm$1.58} & 
  \multicolumn{1}{c|}{\textbf{90.26$\pm$2.43}} &
  \multicolumn{1}{c|}{88.54$\pm$1.94} &
  \multicolumn{1}{c|}{\textbf{89.10$\pm$1.69}} &
  \multicolumn{1}{c|}{90.02$\pm$2.48} &
  \textbf{90.54$\pm$1.37} \\
  \textbf{Amazon Computer} &
  \multicolumn{1}{c|}{82.73$\pm$1.54} &
  \multicolumn{1}{c|}{75.21$\pm$3.56} &
  \multicolumn{1}{c|}{83.38$\pm$2.50} &
  \multicolumn{1}{c|}{67.95$\pm$3.88} & 
  \multicolumn{1}{c|}{\textbf{81.18$\pm$2.53}} &
  \multicolumn{1}{c|}{80.94$\pm$2.33} &
  \multicolumn{1}{c|}{74.01$\pm$3.26} &
  \multicolumn{1}{c|}{74.05$\pm$4.01} &
  \multicolumn{1}{c|}{81.08$\pm$2.54} &
  \textbf{81.56$\pm$2.57} \\ \hline
\end{tabular}
}
\vspace{-3mm}  
\end{table}
\subsection{Study on large-scale graph datasets }We apply experiments to see performance of our framework on large-scale graphs on Arxiv and Products. Results are shown in Table.~\ref{tab:result_large_1} and Table.~\ref{tab:result_large_2}.

\textbf{Results. }We can see that we improve baseline GLNN's performance by about 0.5 to 1.5$\%$ in 7 of 10 total cases via our method, with transductive or inductive setting. It demonstrates that our method really works on large-scale graphs.
\begin{table}[htbp]
\renewcommand\arraystretch{2}
\centering
\caption{Results of teachers, baseline models and KMP for \textbf{trans} setting on large-scale graph datasets. The \textbf{best} performance of student model is highlighted. }
\label{tab:result_large_1}
\resizebox{1\textwidth}{!}{
\begin{tabular}{|c|cccccccccc|} \hline
\multirow{2}{*}{\textbf{Dataset}} &
   \multicolumn{10}{c|}{\textbf{Models}} \\ \cline{2-11}
   &
 \multicolumn{1}{c|}{\textbf{$SAGE$}} &
  \multicolumn{1}{c|}{\textbf{$GCN$}} &
  \multicolumn{1}{c|}{\textbf{$GAT$}} &
  \multicolumn{1}{c|}{\textbf{$MLP$}} &
  \multicolumn{1}{c|}{\textbf{$GLNN(SAGE)$}} &
  \multicolumn{1}{c|}{\textbf{$KMP(SAGE)$}} &
  \multicolumn{1}{c|}{\textbf{$GLNN(GCN)$}} &
  \multicolumn{1}{c|}{\textbf{$KMP(GCN)$}} &
  \multicolumn{1}{c|}{\textbf{$GLNN(GAT)$}} &
  \textbf{$KMP(GAT)$} \\ \hline
   \textbf{OGBN-Arxiv} &
   \multicolumn{1}{c|}{71.54$\pm$0.47} &
   \multicolumn{1}{c|}{69.84$\pm$0.41} &
  \multicolumn{1}{c|}{73.45$\pm$0.29} &
  \multicolumn{1}{c|}{54.12$\pm$2.97} &
  \multicolumn{1}{c|}{\textbf{64.48$\pm$1.29}} & 
  \multicolumn{1}{c|}{64.31$\pm$0.50} &
  \multicolumn{1}{c|}{60.07$\pm$0.18} &
  \multicolumn{1}{c|}{\textbf{60.96$\pm$1.11}} &
  \multicolumn{1}{c|}{63.45$\pm$4.06} &
  \textbf{65.02$\pm$2.23} \\
  \textbf{OGBN-Products} &
  \multicolumn{1}{c|}{78.07$\pm$0.06} &
  \multicolumn{1}{c|}{75.02$\pm$0.23} &
  \multicolumn{1}{c|}{78.35$\pm$0.14} &
  \multicolumn{1}{c|}{60.88$\pm$0.17} &
  \multicolumn{1}{c|}{67.44$\pm$0.60} &
  \multicolumn{1}{c|}{\textbf{68.01$\pm$0.52}} & 
  \multicolumn{1}{c|}{\textbf{66.92$\pm$0.41}} &
  \multicolumn{1}{c|}{66.07$\pm$0.64} &
  \multicolumn{1}{c|}{67.53$\pm$0.33} &
  \textbf{68.20$\pm$0.69} \\ \hline
\end{tabular}
}
\vspace{-3mm}  
\end{table}

\begin{table}[htbp]
\renewcommand\arraystretch{2}
\centering
\caption{Results of teachers, baseline models and KMP for \textbf{induc} setting on large-scale graph datasets. The \textbf{best} performance of student model is highlighted. ('---' means that the GNN shows poor performance and we do not use it as teacher.)}
\label{tab:result_large_2}
\resizebox{1\textwidth}{!}{
\begin{tabular}{|c|cccccccccc|} \hline
\multirow{2}{*}{\textbf{Dataset}} &
   \multicolumn{10}{c|}{\textbf{Models}} \\ \cline{2-11}
   &
  \multicolumn{1}{c|}{\textbf{$SAGE$}} &
  \multicolumn{1}{c|}{\textbf{$GCN$}} &
  \multicolumn{1}{c|}{\textbf{$GAT$}} &
  \multicolumn{1}{c|}{\textbf{$MLP$}} &
  \multicolumn{1}{c|}{\textbf{$GLNN(SAGE)$}} &
  \multicolumn{1}{c|}{\textbf{$KMP(SAGE)$}} &
  \multicolumn{1}{c|}{\textbf{$GLNN(GCN)$}} &
  \multicolumn{1}{c|}{\textbf{$KMP(GCN)$}}
  & 
  \multicolumn{1}{c|}{\textbf{$GLNN(GAT)$}} &
  \textbf{$KMP(GAT)$} \\ \hline
   \textbf{OGBN-Arxiv} &
   \multicolumn{1}{c|}{70.53$\pm$0.58} &
   \multicolumn{1}{c|}{69.79$\pm$0.41} &
  \multicolumn{1}{c|}{---} &
  \multicolumn{1}{c|}{54.46$\pm$2.75} &
  \multicolumn{1}{c|}{58.83$\pm$0.33} & 
  \multicolumn{1}{c|}{\textbf{59.31$\pm$0.22}} &
  \multicolumn{1}{c|}{58.94$\pm$0.71} &
  \multicolumn{1}{c|}{\textbf{60.01$\pm$1.02}} &
  \multicolumn{1}{c|}{\textbf{---}} &
       \textbf{---} \\
  \textbf{OGBN-Products} &
  \multicolumn{1}{c|}{77.25$\pm$0.22} &
  \multicolumn{1}{c|}{74.93$\pm$0.15} &
  \multicolumn{1}{c|}{---} &
  \multicolumn{1}{c|}{61.05$\pm$0.36} &
  \multicolumn{1}{c|}{65.24$\pm$0.39} &
  \multicolumn{1}{c|}{\textbf{65.93$\pm$0.53}} & 
  \multicolumn{1}{c|}{\textbf{64.37$\pm$0.13}} &
  \multicolumn{1}{c|}{63.95$\pm$0.17} &
  \multicolumn{1}{c|}{\textbf{---}} &
  \textbf{---} \\ \hline
\end{tabular}
}
\vspace{-3mm}  
\end{table}

\subsection{Study on trainable reverse kernel } As mentioned in Sec.~\ref{sec:4.3}, we study trainable reverse kernel's performance. We do our study on three benchmark datasets, Cora, Citeseer and Pubmed. Note that we use default hyperparameters for distillation, so we don't really report the 'best' results of our framework here but only a comparison. We show the results in Fig.~\ref{fig:trainable} in Appendix.~\ref{appendix}.
\subsection{Study on Positional Encoding }
Now we consider our KMP's results when utilizing PE as an initial node feature. We test it on the seven datasets mentioned above. We only adopt Laplacian PE, for it's easy to calculate and proved efficient~\cite{ref_article16}. Note that we also introduce a method SA-MLP~\cite{ref_article18} here. It tries to deal with problems of student MLP by simply mapping the adjacency matrix and node features to the same dimension and then concatenating them together. The complete source code of SA-MLP hasn't been provided yet, so we reproduce a training process for it by ourselves. We provide results of SAGE as teacher GNN. Results are listed in Table.~\ref{tab:result_3}.

\textbf{Results. }
\begin{table}[htbp]
\centering
\caption{Results of KMP+PE, KMP and SA-MLP  for both trans and inductive setting. The \textbf{best} performance of the student model is highlighted.}
\label{tab:result_3}
\resizebox{1\textwidth}{!}{
\begin{tabular}{|c|ccc|ccc|} \hline
\multirow{2}{*}{\textbf{Dataset}} &
   \multicolumn{3}{c|}{\textbf{Models(SAGE teacher, trans)}} 
   & \multicolumn{3}{c|}{\textbf{Models(SAGE teacher, induc)}} \\ \cline{2-4} \cline{5-7} &
  \multicolumn{1}{c|}{\textbf{$KMP(ours)$}} &
  \multicolumn{1}{c|}{\textbf{$KMP+PE(ours)$}} &
  \multicolumn{1}{c|}{\textbf{$SA-MLP$}} &
  \multicolumn{1}{c|}{\textbf{$KMP(ours)$}} &
  \multicolumn{1}{c|}{\textbf{$KMP+PE(ours)$}} &
  \multicolumn{1}{c|}{\textbf{$SA-MLP$}} \\ \hline
   \textbf{Cora} &
  \multicolumn{1}{c|}{79.48$\pm$1.48} &
  \multicolumn{1}{c|}{\textbf{80.97$\pm$1.03}} &
  \multicolumn{1}{c|}{79.33$\pm$1.55} &
  \multicolumn{1}{c|}{71.85$\pm$2.40} &
  \multicolumn{1}{c|}{\textbf{71.90$\pm$1.81}} &
  \multicolumn{1}{c|}{71.54$\pm$2.26} \\
  \textbf{Citeseer} &
  \multicolumn{1}{c|}{71.96$\pm$2.25} &
  \multicolumn{1}{c|}{\textbf{72.07$\pm$2.51}} &
  \multicolumn{1}{c|}{71.25$\pm$1.57} &
  \multicolumn{1}{c|}{71.22$\pm$2.04} &
  \multicolumn{1}{c|}{\textbf{71.46$\pm$2.65}} &
  \multicolumn{1}{c|}{70.95$\pm$2.49} \\
  \textbf{Pubmed} &
  \multicolumn{1}{c|}{\textbf{76.80$\pm$2.01}} &
  \multicolumn{1}{c|}{76.58$\pm$2.84} &
  \multicolumn{1}{c|}{76.45$\pm$2.50} &
  \multicolumn{1}{c|}{74.92$\pm$3.03} &
  \multicolumn{1}{c|}{74.71$\pm$2.09} &
  \multicolumn{1}{c|}{\textbf{74.99$\pm$3.74}} \\
  \textbf{Amazon Photo} &
  \multicolumn{1}{c|}{92.27$\pm$1.50} &
  \multicolumn{1}{c|}{\textbf{92.38$\pm$0.82}} &
  \multicolumn{1}{c|}{90.67$\pm$0.93} &
  \multicolumn{1}{c|}{90.26$\pm$2.43} &
  \multicolumn{1}{c|}{\textbf{90.77$\pm$0.84}} &
  \multicolumn{1}{c|}{90.27$\pm$1.44} \\
  \textbf{Amazon Computer} &
  \multicolumn{1}{c|}{84.75$\pm$1.10} &
  \multicolumn{1}{c|}{\textbf{85.10$\pm$1.66}} &
  \multicolumn{1}{c|}{83.38$\pm$2.50} & 
  \multicolumn{1}{c|}{80.94$\pm$2.33} &
  \multicolumn{1}{c|}{\textbf{81.50$\pm$2.58}} &
  \multicolumn{1}{c|}{81.01$\pm$2.89} \\
  \textbf{OGBN-Arxiv} &
  \multicolumn{1}{c|}{64.31$\pm$0.50} &
  \multicolumn{1}{c|}{\textbf{65.32$\pm$0.46}} &
  \multicolumn{1}{c|}{65.03$\pm$0.89} &
  \multicolumn{1}{c|}{59.31$\pm$0.22} &
  \multicolumn{1}{c|}{\textbf{59.47$\pm$0.27}} &
  \multicolumn{1}{c|}{59.26$\pm$0.24} \\
  \textbf{OGBN-Products} &
  \multicolumn{1}{c|}{68.01$\pm$0.52} &
  \multicolumn{1}{c|}{\textbf{68.84$\pm$0.49}} &
  \multicolumn{1}{c|}{67.93$\pm$0.37} &
  \multicolumn{1}{c|}{65.93$\pm$0.53} &
  \multicolumn{1}{c|}{65.41$\pm$0.14} &
  \multicolumn{1}{c|}{\textbf{66.67$\pm$0.30}} \\ \hline
\end{tabular}
}
\vspace{-3mm}  
\end{table}
Now we see that our framework can reach the best performance on all datasets with transductive setting. Moreover, it performs best with inductive setting on 5 datasets. The conclusion is that Positional Encoding can really help KMP work better with little extra consumption for doing calculations. Only a tiny linear layer projection is necessary for change PE into node features(much tinier than SA-MLP). It is a great guide to real-life inference tasks. 
\subsection{Robustness and Sensitiveness study }
\textbf{Dataset. }We do our robustness and sensitiveness study on benchmark datasets: Cora, Citeseer, Pubmed. As they are easy for us to compute student predictions. This is an intuitive guide for large-scale graphs.

\textbf{Problem definition. }We test our framework's robustness by adding noise attack to initial node features $X_{init}$. Consider that a random matrix $X_{rand}$ is added to $X_{init}$ by a certain percentage, and our doubt on robustness is whether KMP can work well with interference of noise features. We choose a percentage in $\{10\%,20\%,30\%,40\%,50\%\}$. Note that we also provide GLNN's performance for comparison. Next, we do sensitiveness test by modifying the constant $\gamma$, mentioned in Sec.~\ref{Sec:4.2}, as our framework is much sensitive to only $\gamma$ revealed by previous experiments. We choose $\gamma \in \{0.1,0.3,0.5,0.7,1,3,10,30\}$. Note that we choose the default hyperparameters and try to get the test curve for comparison but not the best results. 

\textbf{Results. }The results of this robustness study is shown in Fig.~\ref{fig:sensi_1}. (We do not report results on Pubmed, for its test accuracy bitterly declined to very poor values with noise. This can be seen in Appendix.~\ref{appendix:noise}. Our frameworks ability to reduce harm of feature noise is stronger than baseline GLNN by about 2\% in most cases. Next, we lay out results of sensitiveness study in Fig.~\ref{fig:sensi_1}. We discover the fact that our method works stably when $\gamma$ is between 0 and 3, and a sharp decline in test accuracy occurs with $\gamma$ larger than 3. Usually, we choose $\gamma \in (0,1]$ and our framework can work normally. 
\begin{figure}[htbp]
\centering
\subfigure[GCN(Cora)]{
\begin{minipage}[htbp]{0.325\textwidth}
\centering
\includegraphics[width=1.7in]{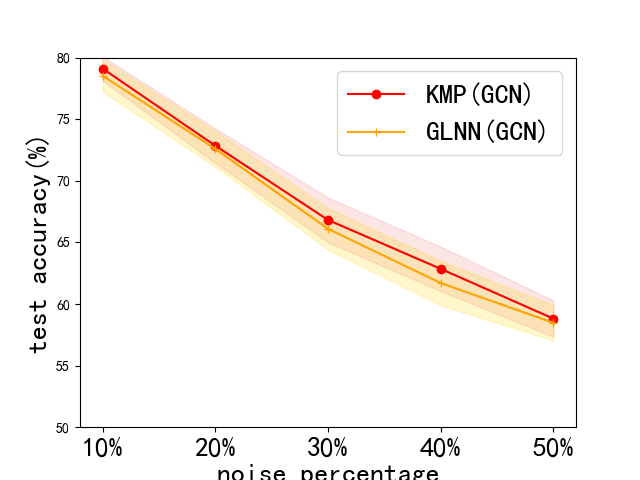}
\end{minipage}%
}%
\subfigure[SAGE(Cora)]{
\begin{minipage}[htbp]{0.325\textwidth}
\centering
\includegraphics[width=1.7in]{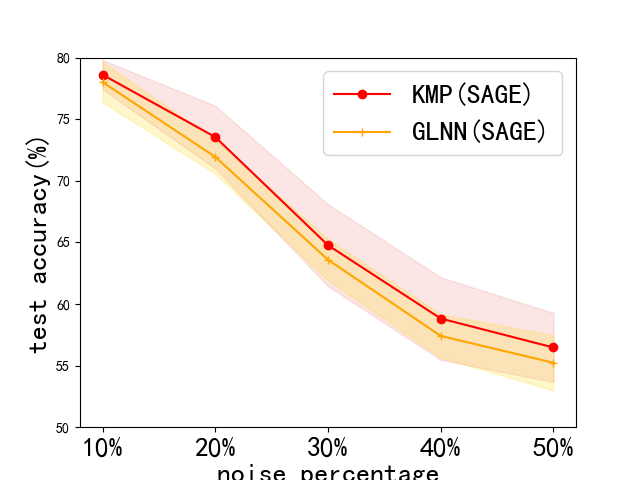}
\end{minipage}%
}%
\subfigure[GAT(Cora)]{
\begin{minipage}[htbp]{0.325\textwidth}
\centering
\includegraphics[width=1.7in]{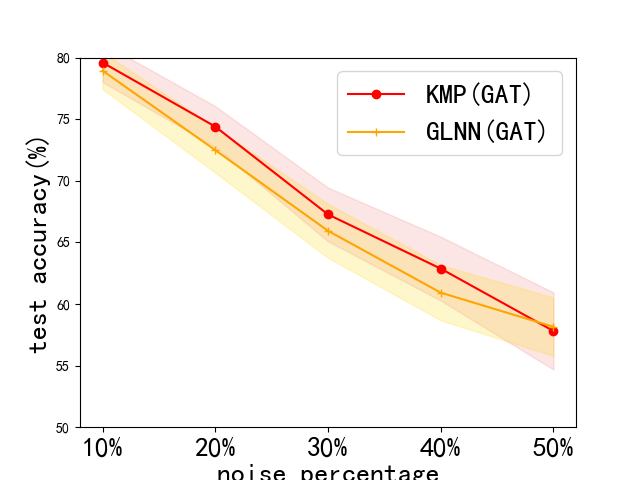}
\end{minipage}
}%
\newline
\subfigure[GCN(Citeseer)]{
\begin{minipage}[htbp]{0.325\textwidth}
\centering
\includegraphics[width=1.7in]{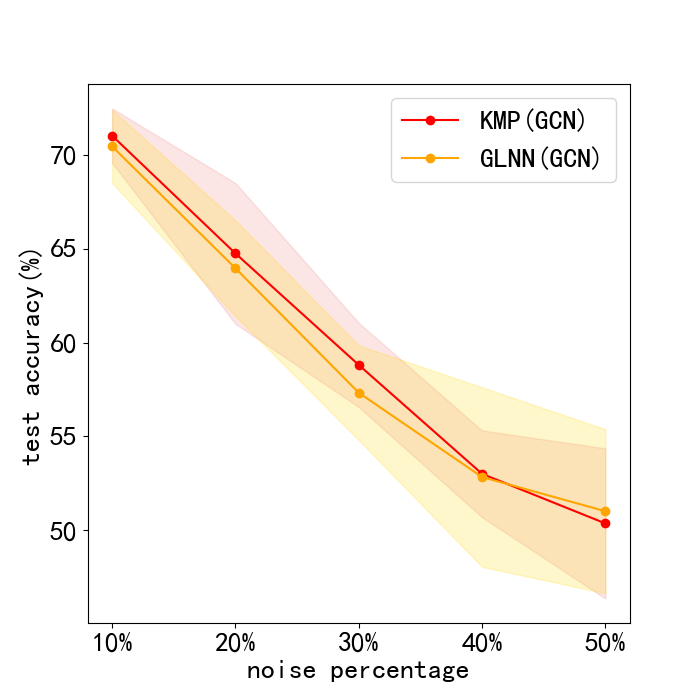}
\end{minipage}%
}%
\subfigure[SAGE(Citeseer)]{
\begin{minipage}[htbp]{0.325\textwidth}
\centering
\includegraphics[width=1.7in]{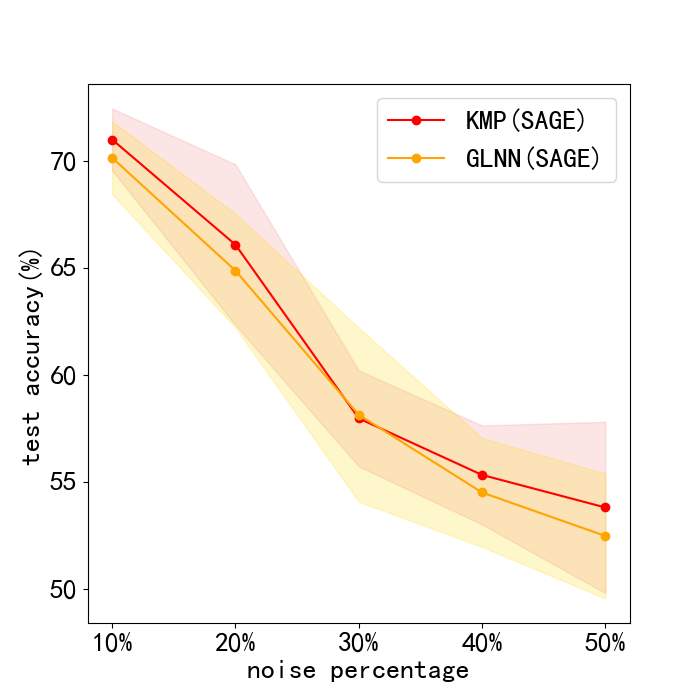}
\end{minipage}%
}%
\subfigure[GAT(Citeseer)]{
\begin{minipage}[htbp]{0.325\textwidth}
\centering
\includegraphics[width=1.7in]{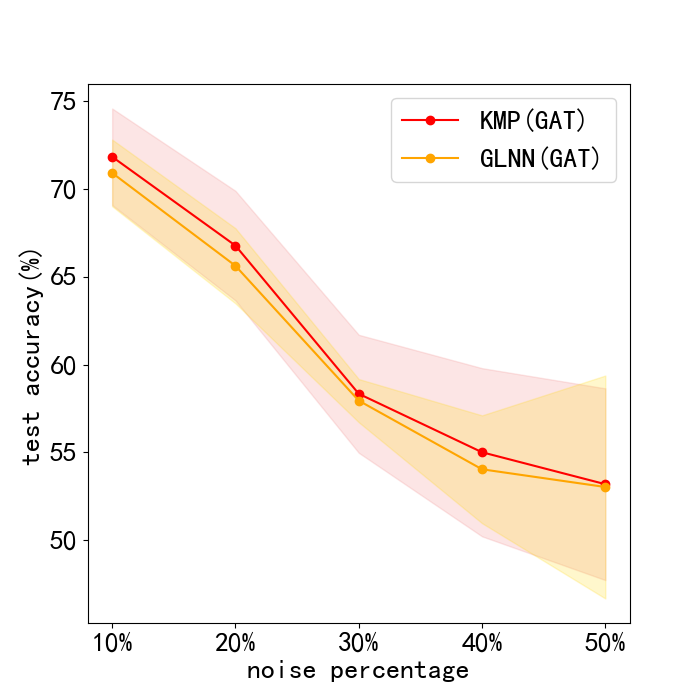}
\end{minipage}
}%
\centering
\caption{Results of Robustness study with different teachers on three datasets. The x-axis shows percentage of feature noise and y-axis means test accuracy(\%).}
\label{fig:robust_1}
\end{figure}

\begin{figure}[htbp]
\centering
\subfigure[Test on Cora]{
\begin{minipage}[htbp]{0.325\textwidth}
\centering
\includegraphics[width=1.7in]{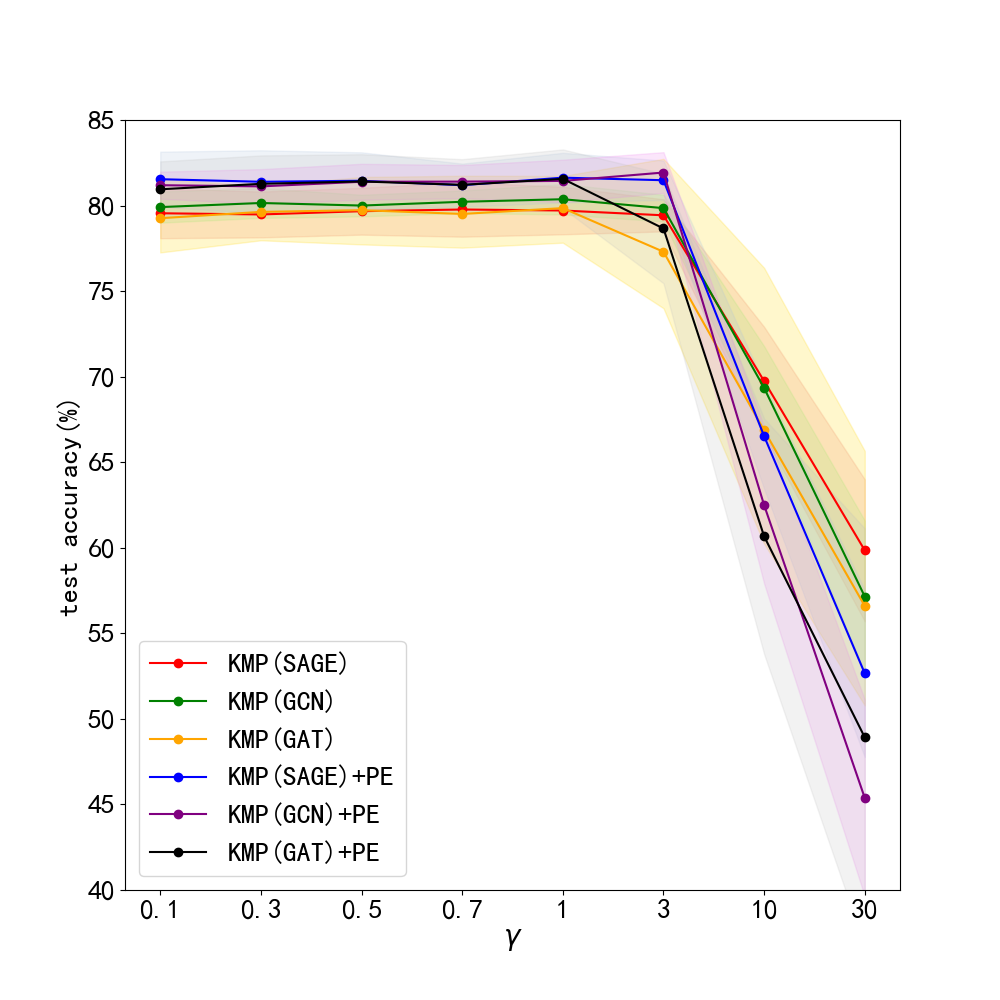}
\end{minipage}%
}%
\subfigure[Test on Citeseer]{
\begin{minipage}[htbp]{0.325\textwidth}
\centering
\includegraphics[width=1.7in]{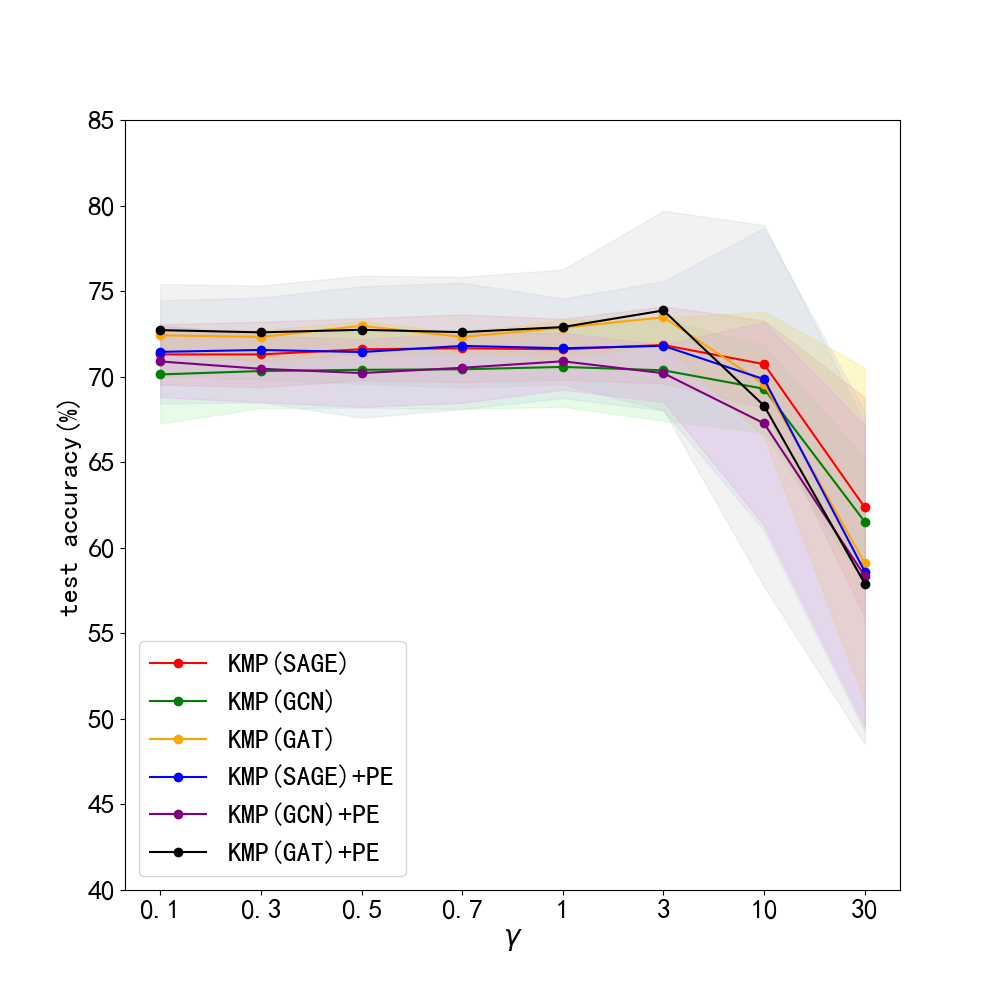}
\end{minipage}%
}%
\subfigure[Test on Pubmed]{
\begin{minipage}[htbp]{0.325\textwidth}
\centering
\includegraphics[width=1.7in]{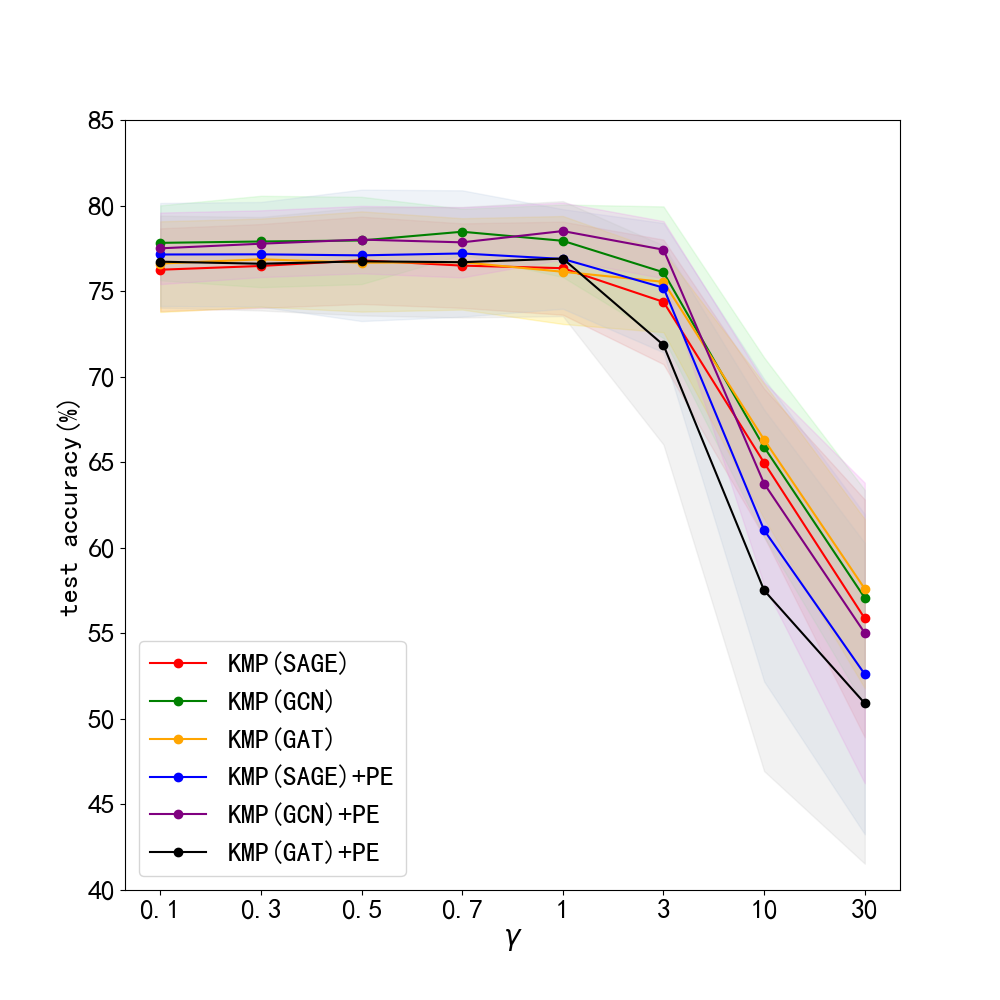}
\end{minipage}
}%
\centering
\caption{Results of sensitiveness study with different teachers on three datasets. Mean and variance of test accuracy reported. We study if changes of $\gamma$ can influence student MLP's performance greatly. The x-axis shows different $\gamma$s and y-axis means test accuracy(\%) with them.}
\label{fig:sensi_1}
\end{figure}

\subsection{Additional study on graph classifaction tasks } Since our framework can achieve good results with inductive setting for node classification, we consider another inductive task: graph classification.

\textbf{Dataset. }We utilize MiniGC~\cite{ref_lncs19} for this additional study. It contains 8 kinds of graphs in total with variable number of nodes. Here we generate different graphs with 10-20 nodes, and choose 240 graphs for training, 60 for validation and 60 for testing. Our target is to train a student MLP excellent enough for classification. 

\textbf{Results. }Our graph classification results are listed in Table.~\ref{tab:result_gcls}. We can see that our KMP increases test accuracy of GLNN by about 3.5$\%$. This means that our approach may also have positive results for larger graph classification datasets. 
\begin{table}[htbp]
\renewcommand\arraystretch{0.75}
\centering
\caption{Results for graph classification on MiniGC dataset. The \textbf{best} performance of student model is highlighted.}
\label{tab:result_gcls}
\resizebox{0.75\textwidth}{!}{
\begin{tabular}{|c|cccc|} \hline
\multirow{2}{*}{\textbf{Dataset}} &
   \multicolumn{4}{c|}{\textbf{Models}} 
   \\ \cline{2-5}  & 
  \multicolumn{1}{c|}{\textbf{$Teacher$}} &
  \multicolumn{1}{c|}{\textbf{$MLP$}} &
  \multicolumn{1}{c|}{\textbf{$GLNN$}} &
  \multicolumn{1}{c|}{\textbf{$KMP$}} 
   \\ \hline
   \textbf{MiniGC} &
  \multicolumn{1}{c|}{91.90$\pm$3.47} &
  \multicolumn{1}{c|}{66.35$\pm$5.36} &
  \multicolumn{1}{c|}{75.44$\pm$4.69} &
  \multicolumn{1}{c|}{\textbf{78.95$\pm$5.01}} 
 \\ \hline
\end{tabular}
}
\vspace{-3mm}  
\end{table}
\section{Conclusion and Future work}
This paper formalizes the two main problems of transferring knowledge from teacher GNN to student MLP. Hence we propose a new framework for teaching student via knowledge distillation on graphs. In detail, we utilize postional encoding as additional initial node feature to solve the problem of structural information loss. What's more, we introduce heat kernels in GNNs and use it for hidden layer distillation between GNNs and MLPs. Experiment results reveal us that our framework can reach the best result in almost all cases, and it obtains better robustness than baseline models. 

We would like to point out several future directions: During training in our current work, the dimension of student MLP's hidden layers must match that of teacher GNN's hidden layers. It makes us hard to use a wider MLP as student without changing hidden layer dimensions of teacher model. Also, our current work can only introduce most frequently used GNNs as teachers, and they may perform badly in certain tasks. And that will cause to poor performance of student MLP. We will focus on more flexible methods for hidden layer distillation and adopt more sophisticated models as teachers, for example, Graph Transformer, etc. 
\section{Ethics}
Our paper does not involve any ethical or moral issues. It is purely focused on proposing a methodology to solve two major problems of student MLP. Our experiments mainly aim to demonstrate the effectiveness and robustness of our proposed framework. Therefore, there is no need for ethical or moral considerations in our research. We ensure that our research is conducted in compliance with ethical guidelines and regulations.

\bibliographystyle{splncs04}
\bibliography{References.bib}

\clearpage
\appendix
\appendix
\label{appendix}
\section{Appendix}

\subsection{Dataset details }The seven used dataset for main experiments are listed here. 
\begin{table}[htbp]
    \caption{Details of the seven used datasets}
    \label{tab:CPF_detail}
    \centering
    \begin{tabular}{|c|c|c|c|c|} \hline
     \textbf{Dataset} & 
     \multicolumn{1}{c|}{Nodes}  & 
     \multicolumn{1}{c|}{Edges} &
     \multicolumn{1}{c|}{Feature dim} &
     \multicolumn{1}{c|}{Classes} \\ \hline  
     \textbf{Cora} & 
     \multicolumn{1}{c|}{2708}  & 
     \multicolumn{1}{c|}{5429} &
     \multicolumn{1}{c|}{1433} &
    7 \\
    \textbf{Citeseer} & 
     \multicolumn{1}{c|}{3327}  & 
     \multicolumn{1}{c|}{4732} &
     \multicolumn{1}{c|}{3703} &
    6 \\
     \textbf{Pubmed} & 
     \multicolumn{1}{c|}{19717}  & 
     \multicolumn{1}{c|}{44324} &
     \multicolumn{1}{c|}{500} &
    3 \\
     \textbf{Amazon Photo} & 
     \multicolumn{1}{c|}{7487}  & 
     \multicolumn{1}{c|}{119043} &
     \multicolumn{1}{c|}{745} &
    8 \\
     \textbf{Amazon Computer} & 
     \multicolumn{1}{c|}{13381}  & 
     \multicolumn{1}{c|}{245778} &
     \multicolumn{1}{c|}{767} &
    10 \\
    \textbf{OGBN-Arxiv} & 
     \multicolumn{1}{c|}{169343}  & 
     \multicolumn{1}{c|}{1166243} &
     \multicolumn{1}{c|}{128} &
      40 \\
      \textbf{OGBN-Products} & 
     \multicolumn{1}{c|}{2449029}  & 
     \multicolumn{1}{c|}{61859140} &
     \multicolumn{1}{c|}{100} &
      47 \\ \hline
    \end{tabular}
\end{table}

\subsection{Settings mentioned in Sec.~\ref{Sec:5.4}} \label{appendix:settings}
We follows the settings for Cora, Citeseer, Pubmed, Amazon Photo and Amazon Computer as:
\begin{table}[htbp]
    \caption{Experiment settings for the 5 datasets.}
    \label{tab:settings}
    \centering
    \begin{tabular}{|c|c|c|c|} \hline
        & 
     \multicolumn{1}{c|}{GCN} &
     \multicolumn{1}{c|}{SAGE} &
     \multicolumn{1}{c|}{GAT} \\ \hline  
      number of layers & 
     \multicolumn{1}{c|}{2}  & 
     \multicolumn{1}{c|}{2} &
     \multicolumn{1}{c|}{2} \\
     hidden layer dim & 
     \multicolumn{1}{c|}{64}  & 
     \multicolumn{1}{c|}{128} &
     \multicolumn{1}{c|}{128}  \\
     weight decay &
     \multicolumn{1}{c|}{0.001}  & 
     \multicolumn{1}{c|}{5e-4} &
     \multicolumn{1}{c|}{0.01}  \\
     dropout ratio &
      \multicolumn{1}{c|}{0.8}  & 
     \multicolumn{1}{c|}{0} &
     \multicolumn{1}{c|}{0.6}  \\
     attention dropout ratio &
      \multicolumn{1}{c|}{---}  & 
     \multicolumn{1}{c|}{---} &
     \multicolumn{1}{c|}{0.6} \\
     fan\_out &
      \multicolumn{1}{c|}{---}  & 
     \multicolumn{1}{c|}{5,5} &
     \multicolumn{1}{c|}{---} \\
     number of heads &
      \multicolumn{1}{c|}{---}  & 
     \multicolumn{1}{c|}{---} &
     \multicolumn{1}{c|}{3} \\
     norm type(if used) &
      \multicolumn{1}{c|}{batch}  & 
     \multicolumn{1}{c|}{batch} &
     \multicolumn{1}{c|}{batch} \\ \hline
    \end{tabular}
\end{table}

\subsection{Grid search of hyperparameters.} \label{appendix:grid_search}
We do grid search for the best hyperparameters for main experiments in Table.~\ref{tab:grid}. We choose hyperparameters from the listed values. Note that $\gamma$ influences student MLP's performance most. 
\begin{table}[htbp]
    \caption{Grid Search of hyperparameters. }
    \label{tab:grid}
    \centering
    \begin{tabular}{|c|c|} \hline
    $\gamma$ &
    \multicolumn{1}{c|}{0.1,0.3,0.5,0.7,0.9,1,3,10,30}  \\ \hline 
    temperature($t$)
    & \multicolumn{1}{c|}{0.25,0.5,1,2,4,10}
    \\ \hline
    $\theta$(percentage of hard labels) &
    \multicolumn{1}{c|}{0,0.2,0.4,0.6,0.8,1} \\ \hline
    learning rate &
    \multicolumn{1}{c|}{0.01,5e-3,1e-3,5e-4,1e-4} \\ \hline
    \end{tabular}
\end{table}
\newline \noindent We also notice that a higher $\gamma$ may bring better prediction results when making use of PE.

\subsection{Results of trainable kernels }
\begin{figure}[htbp]
\centering
\subfigure[Trainable kernel for Cora. ]{
\begin{minipage}[t]{0.5\textwidth}
\centering
\includegraphics[width=2.4in]{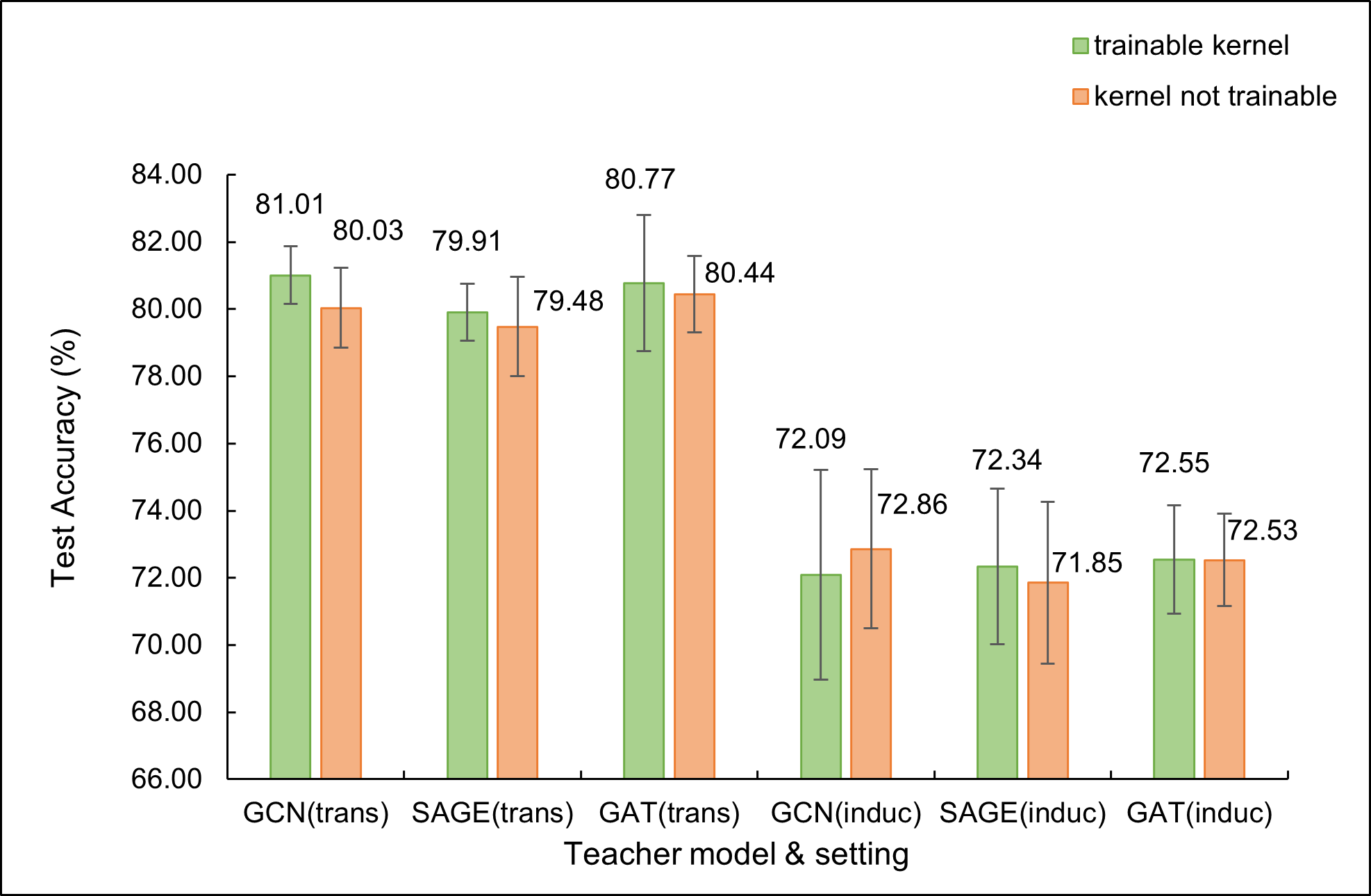}
\end{minipage}%
}%
\subfigure[Trainable kernel for Citeseer. ]{
\begin{minipage}[t]{0.5\textwidth}
\centering
\includegraphics[width=2.4in]{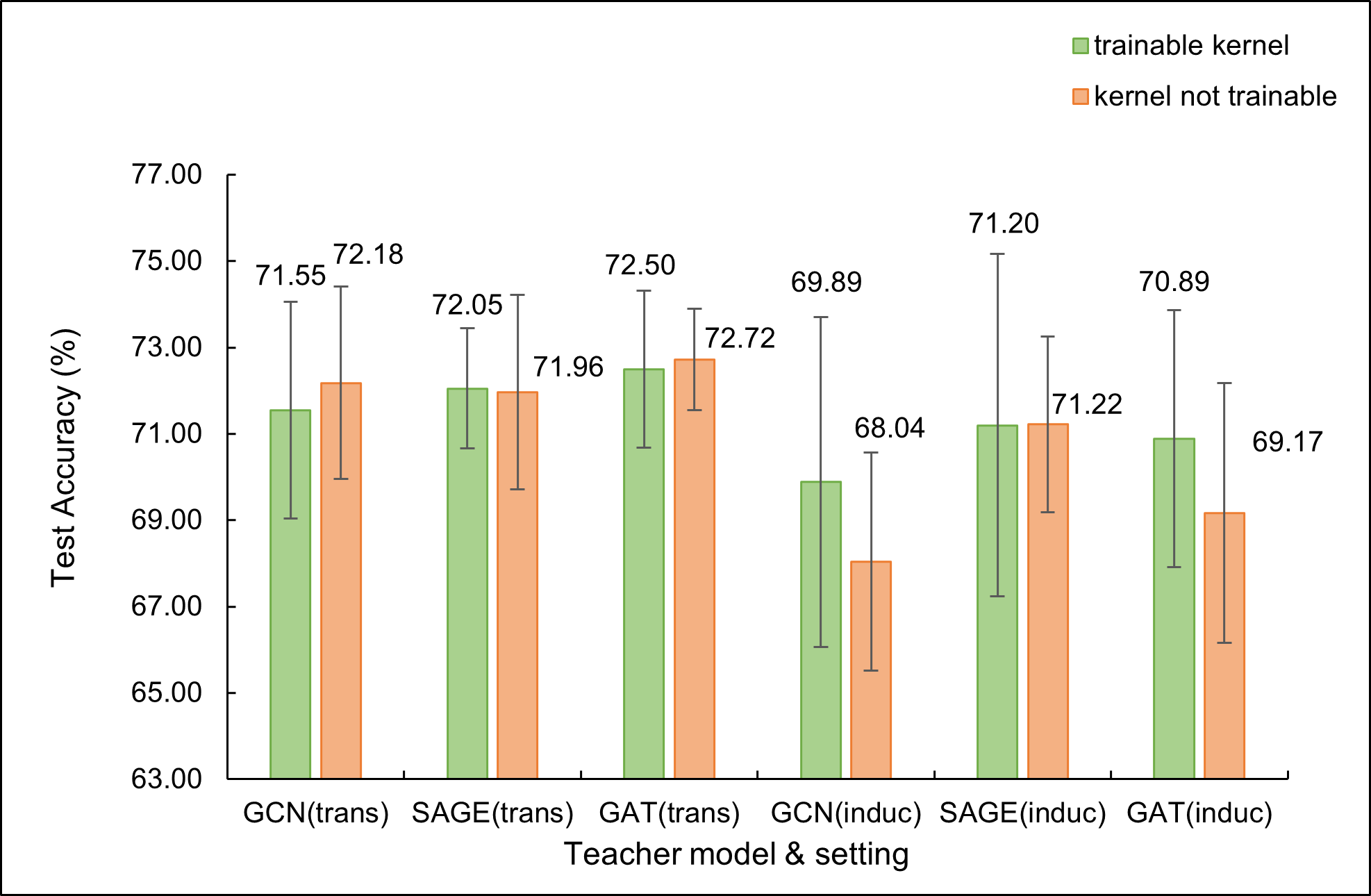}
\end{minipage}%
}%
\newline
\subfigure[Trainable kernel for Pubmed. ]{
\begin{minipage}[t]{0.5\textwidth}
\centering
\includegraphics[width=2.4in]{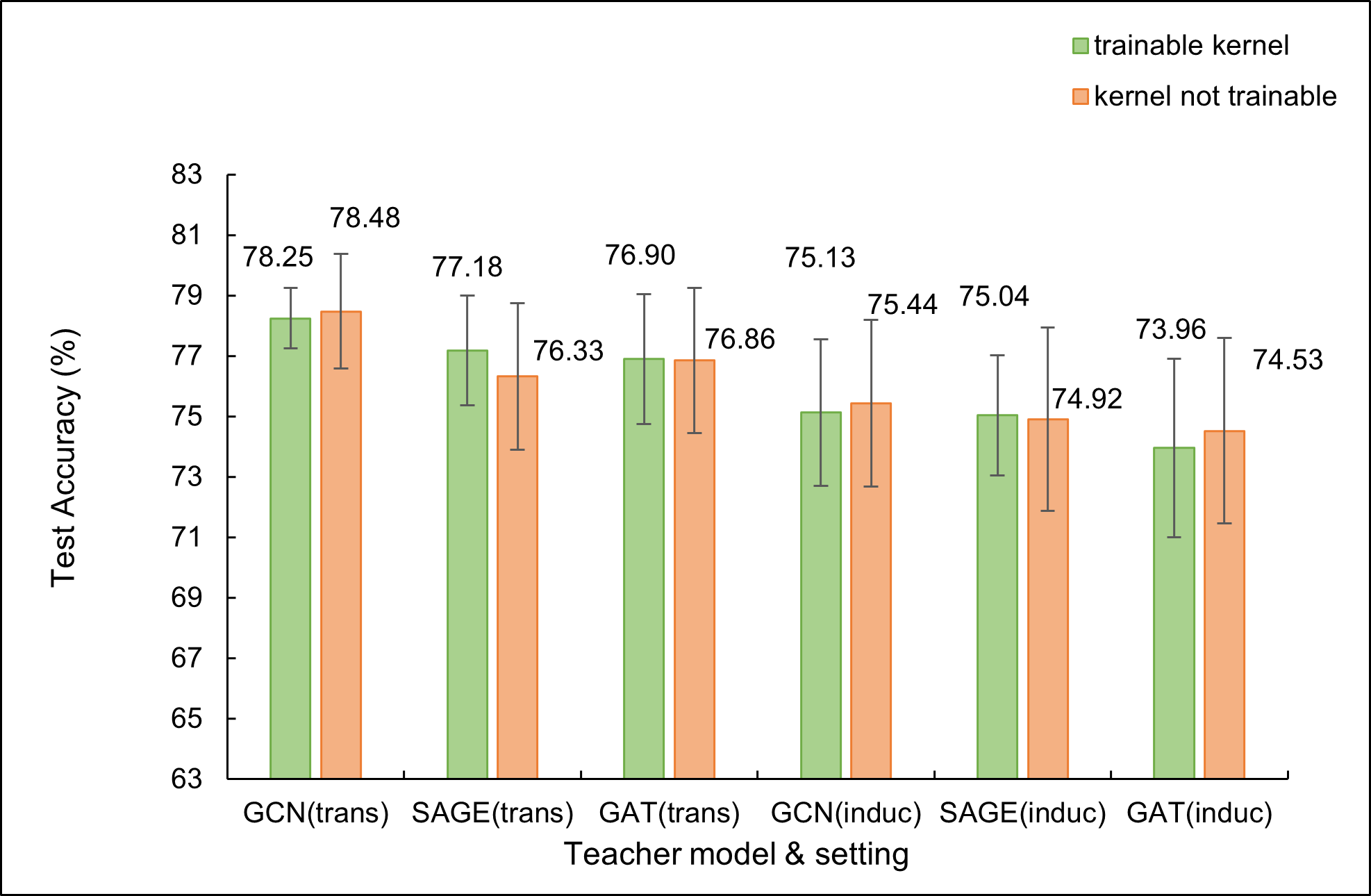}
\end{minipage}
}%
\centering
\caption{Results of trainable kernel on three benchmark datasets, compared to kernels without self training }
\label{fig:trainable}
\end{figure}
As shown in Fig.~\ref{fig:trainable}, trainable kernels can improve student MLP's performance in many cases, especially GCN, GAT as teacher for Citeseer with inductive setting. Nevertheless, training student MLP together with kernel training can converge more slowly, for a two-stage optimization is introduced. Thus, it demonstrates us to choose trainable kernel in proper situations. 

\subsection{Perfomance on Pubmed with noise.} \label{appendix:noise}
When adding noise to node features of the Pubmed dataset, we see a very poor performance as Table.~\ref{tab:pub_noise} shows:
\begin{table}[htbp]
    \caption{Results on Pubmed(with feature noise).}
    \label{tab:pub_noise}
    \centering
    \begin{tabular}{|c|c|c|} \hline
    Percentage of noise & 
     GLNN(SAGE teacher) & 
     KMP(SAGE teacher) \\ \hline
      10\% & 
     \multicolumn{1}{c|}{47.70$\pm$2.91}  & 
     \multicolumn{1}{c|}{48.91$\pm$2.44} \\
      20\% & 
     \multicolumn{1}{c|}{40.75$\pm$2.29}  & 
     \multicolumn{1}{c|}{41.60$\pm$2.47} \\
     30\% & 
     \multicolumn{1}{c|}{37.08$\pm$3.96}  & 
     \multicolumn{1}{c|}{37.41$\pm$4.65} \\
     40\% & 
     \multicolumn{1}{c|}{33.96$\pm$5.23}  & 
     \multicolumn{1}{c|}{33.16$\pm$5.88} \\
      50\% & 
     \multicolumn{1}{c|}{\textbf{---}}  & 
     \multicolumn{1}{c|}{\textbf{---}} 
       \\ \hline
    \end{tabular}
\end{table}

\end{document}